\documentclass[conference]{IEEEtran}

\usepackage{graphicx}
\usepackage{amsmath}
\usepackage{amsfonts}

\ifCLASSINFOpdf
\else
\fi
\DeclareMathOperator*{\argmax}{\arg\!\max}

\begin{document}
%
\title{Cross-Device Tracking: \\ Matching Devices and Cookies}


\author{\IEEEauthorblockN{Roberto D\'iaz-Moralesl\IEEEauthorrefmark{1}\IEEEauthorrefmark{2}}
\IEEEauthorblockA{\IEEEauthorrefmark{1}IDI Department,Treelogic\\ Parque Tecnol\'ogico de Asturias, parcela 30, 33428, Spain\\ Email: roberto.diaz@treelogic.com}
\IEEEauthorblockA{\IEEEauthorrefmark{2}DTSC, University Carlos III de Madrid, Spain\\
Email: rdiazm@tsc.uc3m.es}}



%


\maketitle

\begin{abstract}
The number of computers, tablets and smartphones is increasing rapidly, which entails the ownership and use of multiple devices to perform online tasks. As people move across devices to complete these tasks, their identities becomes fragmented.

Understanding the usage and transition between those devices is essential to develop efficient applications in a multi-device world. In this paper we present a solution to deal with the cross-device identification of users based on semi-supervised machine learning methods to identify which cookies belong to an individual using a device. The method proposed in this paper scored third in the ICDM 2015 Drawbridge Cross-Device Connections challenge proving its good performance.
\end{abstract}


%
\IEEEpeerreviewmaketitle

\section{Introduction}
With the rapid adoption of multiple devices  (such as desktop computers, laptops, smartphones and tablets) the consumptions habits have changed \cite{comscore}. People have access to different devices almost anytime anywhere \cite{times}, and they employ multiple devices to complete their online objectives. For these reasons, the data used to understand their behaviors are fragmented and the identification of users becomes challenging.

Some works \cite{compo1}\cite{compo2}\cite{compo3}\cite{compo4}\cite{compo5}\cite{compo6} have been dedicated to study and understand the behavior of the users and how they perform different tasks through different devices.

The goal of cross-device targeting or tracking is to know if the person using computer X is the same one that uses mobile phone Y and tablet Z. This is an important emerging technology challenge and a hot topic right now because this information could be especially valuable for marketers, due to the possibility of serving targeted advertising to consumers regardless of the device that they are using. Empirically, marketing campaigns tailored for a particular user have proved themselves to be much more effective than general strategies based on the device that is being used. 

Currently, some big companies like Facebook or Google offer this kind of services \cite{Facebook}\cite{Google}, but they need the user to be signed in to their websites and apps. This requirement is not met in several cases.

The usual way to tackle this problem is by using deterministic information and exact match rules \cite{rules} (credit cards numbers, email addresses, mobile phone numbers, home addresses,...). These solutions can not be used for all users or platforms.

Without personal information about the users, cross-device tracking is a complicated process that involves the building of predictive models that have to process many different signals.

In this paper, to deal with this problem, we make use of relational information about cookies, devices, as well as other information like IP addresses to build a model able to predict which cookies belong to a user handling a device by employing semi-supervised machine learning techniques.

The rest of the paper is organized as follows. In Section 2, we talk about the dataset and we briefly describe the problem. Section 3 presents the algorithm and the training procedure. The experimental results are presented in section 4. In section 5, we provide some conclusions and further work. Finally, we have included two appendices, the first one contains information about the features used for this task and in the second a detailed description of the database schema provided for the challenge.

\section{The Challenge}

\subsection{The competition}

This challenge, organized by Drawbridge \cite{drawbridge} and hosted by Kaggle \cite{kaggle}, took place from June 1st 2015 to August 24th 2015 and it brought together 340 teams.

\subsection{The goal}

Users are likely to have multiple identifiers across different domains, including mobile phones, tablets and computing devices. Those identifiers can illustrate common behaviors, to a greater or lesser extent, because they often belong to the same user. Usually deterministic identifiers like names, phone numbers or email addresses are used to group these identifiers.

In this challenge the goal was to infer the identifiers belonging to the same user by learning  which cookies belong to an individual using a device. Relational information about users, devices, and cookies was  provided, as well as other information on IP addresses and behavior. See appendix Dataset and the official challenge webpage \cite{official} for more information about the dataset provided in this challenge.

\subsection{Evaluation metric}

The objective of the challenge was to get the classifier with the highest $F_{0.5}$ score. This score, commonly used in information retrieval, measures the accuracy using the precision $p$ and recall $r$.

Precision is the ratio of true positives ($tp$) to all predicted positives, true positives and false positives ($tp + fp$). Recall is the ratio of true positives to all actual positives, true positives and false negatives ($tp + fn$). 

\begin{align}
&{\cal F}_{\beta}=(1+\beta^2)\frac{pr}{\beta^2p+r}, \nonumber\\
&p=\frac{tp}{tp+fp}, \nonumber\\
&r=\frac{tp}{tp+fn} \nonumber\\
\end{align}

By using $\beta=0.5$ the score weighs precision higher than recall. The score is formed by averaging the individual ${\cal F}_{0.5}$ scores for each device in the test set.

\section{The Algorithm}
\label{alg}

\subsection{Preprocessing}

At the initial stage, we iterate over the list of cookies looking for other cookies with the same handle.

Then, for every pair of cookies with the same handle, if one of them doesn't appear in an IP address that the other cookie appears, we include all the information about this IP address in the cookie.

\subsection{Initial selection of candidates}
\label{candidates}

It is not possible to create a training set containing every combination of devices and cookies due to the high number of them. In order to reduce the initial complexity of the problem and to create a more manageable dataset, some basic rules have been created to obtain an initial reduced set of eligible cookies for every device.

The rules are based on the IP addresses that both device and cookie have in common and how frequent they are in other devices and cookies. Table \ref{RulesTable} summarizes the list of rules created to select the initial candidates.

{
\begin{table}[ht!]
\small
\caption{Rules to select the initial training and test set instances}
\begin{tabular}{p{0.45\textwidth}}
\hline
\\
For every device in the training set we apply the following rules to select the eligible cookies:\\
\\
\hline
\underline{\textbf{Rule 1:}}\\
\\
We create a set that contains the device's IP addresses that appear in less than ten devices and less than twenty cookies. The initial list of candidates is every cookie with known handle that appears in any of these IP addresses.\\
\\
\hline
\underline{\textbf{Rule 2:}}\\
\\If the previous rule returned an empty set of candidates:\\
We create a set that contains the device's IP addresses that appear in less than twenty five devices and less than fifty cookies. The initial list of candidates is every with known handle cookie that appears in any of these IP addresses.\\
\\
\hline
\underline{\textbf{Rule 3:}}\\
\\If the previous rule returned an empty set of candidates:\\
We create a set that contains the device's IP addresses. The initial list of candidates is every cookie with known handle that appears in any of these IP addresses.\\
\\
\hline
\underline{\textbf{Rule 4:}}\\
\\If the previous rule returned an empty set of candidates:\\
We create a set that contains the device's IP addresses. The initial list of candidates is every cookie that appears in any of these IP addresses.\\
\\
\hline
\underline{\textbf{Rule 5:}}\\
\\If a cookie has the same handle than any of the candidates then this cookie is a candidate too.\\
\\
\hline
\end{tabular}
\label{RulesTable}
\end{table}
}

This selection of candidates is very effective in an initial stage, as it reduces the size of the training set to a number of 3664022 device/cookie and to a number 1505453 pairs in the test set. In 98.3\% of the devices, the set of candidates contains the device's cookies.

\subsection{The features}
\label{feat}

Every sample in the training and test set represents a device/eligible cookie pair obtained by the previous step and is composed by a total of 67 features that contains information about the device (Operating System (OS), Country, ...), the cookie (Cookie Browser Version, Cookie Computer OS,...) and the relation between them (number of IP addresses shared by both device and cookie, number of other cookies with the same handle than this cookie,...). 

Appendix dataset shows a detailed description of every feature contained in a training sample.

\subsection{Supervised Learning}
\label{SL}
To create the classifier, we have selected a Regularized Boosted Trees algorithm. Boosting techniques build models using the information of weak predictors, typically decision trees.

We have used logistic regression for binary classification as the learning objective in order to obtain a probabilistic classification that models the probability of the eligible cookie belonging to the device.

This algorithm partitions the data into clusters and uses the cluster index as new features to further minimize the objective function. To create these clusters the algorithm recursively partitions data instances by growing a binary tree. The software that we used was XGBoost \cite{XGBoost}, an open source C++ implementation that utilizes OpenMP to perform automatic parallel computation on a multi-threaded CPU to speedup the training procedure. It has proven its efficiency in many challenges \cite{XGB}\cite{XGR}\cite{NER}.

The parameters of the algorithm have been obtained using a 10 fold cross validation approach:

\begin{itemize}
\item Round for boosting: 200 
\item Maximum depth: 10
\item Subsampling ratio: 1.0
\item Minimum loss reduction for leave partition: 4.0
\item Step Size: 0.1
\item Gamma: 5.0
\end{itemize}

\subsection{Bagging}
\label{B}
Bootstrap aggregating \cite{bagging}, also called bagging, is a technique that allows us to improve the stability of machine learning algorithms. It reduces the variance, and it helps avoid overfitting, resulting in an accuracy improvement.

Having a training set $\mathbf{X}$ with n samples it is possible to generate different training sets sampling from $\mathbf{X}$, then a model for every dataset can be built and all of them can be combined by averaging the output. Experimental and theoretical results suggest that this technique can push unstable procedures towards optimality.

For this problem, we have used eight baggers in the training procedure.

\subsection{Semi-Supervised Learning}
\label{SSL}
Semi-supervised learning is a class of supervised learning that also makes use of unlabeled data. In our case we make use of the data contained in the test set. After scoring the eligible cookies, if the highest score is close to 1 and the second highest score is close to 0, is very likely that the first cookie belongs to the device.

\begin{figure}[h!]
\centering
\includegraphics[width=3.5in]{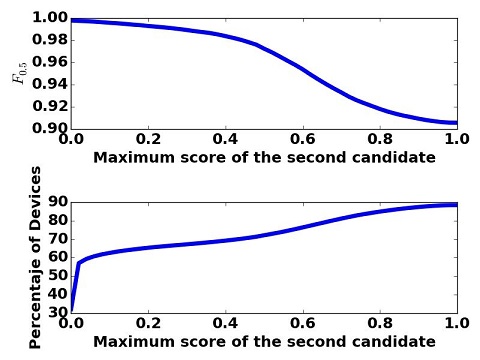}
\caption{$F_{0.5}$ score and percentage of devices when only devices whose second candidate scores less than a certain level are taken into account}
\label{fig_accepted}
\end{figure}

Figure \ref{fig_accepted} shows the $F_{0.5}$ score and the percentage of devices that we are using in its evaluation when we take into account only the devices whose second candidate scores less than a certain threshold. For devices where the second candidate scores less than 0.1 the average $F_{0.5}$ score is higher than 0.99 and 62\% of the devices satisfy that condition.

In our case, we have taken the devices of the test set where the first candidate scores higher than 0.4 and the second candidate scores less than 0.05 and we have considered them to recalculate some features of the training set and retrain the algorithm again (see the sets $\mathcal{O}$, $\mathcal{I}_O$ and $\mathcal{P}_O$ in the appendix Features).

\subsection{Post-processing}
\label{PP}
After obtaining the predictions of the Regularized Boosted Trees, we select the device's cookies using the procedure described in table \ref{SelectingTable}.\\

 {
\begin{table}[ht!]
\small
\caption{Selecting the cookies}
\begin{tabular}{p{0.45\textwidth}}
\hline
\\
For every Device we have a set of $n$ eligible cookies, every candidate has a score $\hat{y}$ obtained by the Regularized Boosted Trees:\\
\\
$\{(\hat{y}_1,C_1),..,(\hat{y}_n,C_n)\}$\\
\\
\hline
\underline{\textbf{Step 1:}}\\
\\
We take the winning candidate:\\
\\
 $(\hat{y}_k,C_k)$, where $k=\underset{i}\argmax(\hat{y}_i)$\\
\\
If $\hat{y}_k > Theshold \to$ Jump to Step 3\\
\\
\hline
\underline{\textbf{Step 2:}}\\
\\
The initial selection of candidates described in section \ref{candidates} did not find a candidate with enough likelihood. We choose a new set of eligible cookies selecting every cookie that shares an IP address with the device and we score them using the classifier: \\
\\
$\{(\hat{y}_1,C_1),..,(\hat{y}_m,C_m)\}$\\
\\
After that, we select the cookie that achieves the highest score:\\
\\
 $(\hat{y}_k,C_k)$, where $k=\underset{i}\argmax(\hat{y}_i)$\\
\\
\hline
\underline{\textbf{Step 3:}}\\
\\
We label $C_k$ as one of the device's cookies. If there are other cookies with the same handle than $C_k$ we label them too.\\
\\
\hline
\underline{\textbf{Step 4:}}\\
\\
We sort the candidates in descending order by the score they have reached and we iterate over them.

If $\hat{y}_i > \hat{y}_k * AccessParameter$ We label $C_i$ as one of the device's cookies. If there are other cookies with the same handle than $C_i$ we label them too.\\

There are different values of $AccessParameter$ attending to:
\begin{itemize}
\item The number of cookies already labeled as device's cookies.
\item The number of other cookies with the same handle than $C_i$
\item The handle of $C_k$ is known or not.
\item The handle of $C_i$ is known or not.
\end{itemize}
\\
\hline
\\
The value of the $threshold$ in step 1 and the different values of $AccessParameter$ in step 4 have been obtained using 10 fold cross validation technique.

It must be taken into account that the training of the training of the Regularized Boosted Trees has been done in a previous step and it is not necessary to retrain it to validate every value of these parameters.
\\
\\
\hline
\end{tabular}
\label{SelectingTable}
\end{table}
}

\section{Results}

The score of this competition was evaluated using a test set of 61156 devices. During the challenge, the leaderboard was calculated on 30\% of the test devices (called public leaderboard). After the competition ended, the final result was known and was based on the other 70\% of the test data (called private leaderboard).

Table \ref{scores} shows the scores in both public and private leaderboard of the different procedures described in section \ref{alg}.

\begin{table}[]
\centering
\caption{Scores in public and private leaderboard}
\label{scores}
\begin{tabular}{lcc}
 \multicolumn{3}{l}{Sel = Initial selection of candidates (\ref{candidates})}\\
 \multicolumn{3}{l}{SL = Supervised Learning (\ref{SL})}\\
 \multicolumn{3}{l}{B = Bagging (\ref{B})}\\
 \multicolumn{3}{l}{SSL = Semi-Supervised Learning (\ref{SSL})}\\
 \multicolumn{3}{l}{PP = Post Processing (\ref{PP})}\\
\\
Procedures & Public Leaderboard & Private Leaderboard  \\
 Sel & 0.498 & 0.5  \\
 IL + SL & 0.872 & 0.875   \\
 Sel + SL + B & 0.874 & 0.876   \\
 Sel + SSL +B + PP & 0.878 & 0.88   
\end{tabular}
\end{table}

Using just the initial selection of candidates described in section \ref{candidates} the $F_{0.5}$ score is 0.5.

When we use the supervised learning procedure described  in section \ref{SL} and we select the cookie with the highest score and other cookies with the same handle than this one, the $F_{0.5}$ is 0.875.

Including the bagging procedure the $F_{0.5}$ score increases to $0.876$.

The full procedure, (Selection of candidates + Semi-Supervised Learning + Bagging + Post Processing) reached an $F_{0.5}$ score of 0.88 in the private leaderboard finishing in third position.

\section{Conclusions}

In this paper we have presented the design, implementation and evaluation of a way to match devices and cookies to deal with the Cross-Device tracking problem.

The source code is publicly available at https://github.com/RobeDM/ICDM2015/ .

This procedure has proved its good performance in the ICDM 2015 Drawbridge Cross-Device Challenge finishing in third place among the solutions of  340 teams and reaching an $F_{0.5}$ score of 0.88.

After these initial results, we can suggest some future research lines:

\begin{itemize}
\item The dataset contained information about the device, cookies and IP addresses. As a first research line we propose to explore other features (for example temporal information) that could be useful for this task.  
\item In our solution we have used one single algorithm. An ensemble of different algorithms could be an interesting option because these techniques obtain better results if there is diversity among the classifiers.
\end{itemize}

\appendix[Dataset]
\label{appen}

The goal was to determine which cookies belong to an individual using a device. You were provided with relational information about users (drawbridge\_handle column), devices, cookies, as well as other information on IP addresses and behavior. The dataset contains fabricated non personally identifiable IDs and is detailed in this section.

\subsection{Device information tables}

This table contains basic information about the devices. It is divided into train and test parts. This is the schema:

\begin{table}[h!]
\begin{tabular}{|p{2.2cm}|p{6cm}|}
\hline
Feature & Description \\\hline
Drawbridge Handle & Identifies the person behind a device. If device and cookie belong to the same person, they will have the same handle. It is unknown for the test subset.\\
Device ID & Identifier of each device.\\
Device type & Categorical feature that indicates the kind of mobile phone, tablet or computer.\\
Device Os & The version of the operating system.\\
Device Country & Which country the device belongs to.\\
Annonymous\_c0 & Anonymous boolean feature.\\
Annonymous\_c1 & Anonymous categorical feature.\\
Annonymous\_c2 & Anonymous categorical feature.\\
Annonymous\_5 & Anonymous feature to describe device.\\
Annonymous\_6 & Anonymous feature to describe device.\\
Annonymous\_7 & Anonymous feature to describe device.\\\hline
\end{tabular}
\end{table}

\subsection{Cookie information tables}

A high-level summary information regarding the cookie:

\begin{table}[h!]
\begin{tabular}{|p{2.2cm}|p{6cm}|}
\hline
Feature & Description \\\hline
Drawbridge Handle & Identifies the person behind a cookie.\\
Cookie ID & Identifier of each cookie.\\
Computer OS & Cookie computer operation system.\\
Browser Version & Cookie browser version.\\
Device Country & Which country the cookie belongs to.\\
Annonymous\_c0 & Anonymous boolean feature.\\
Annonymous\_c1 & Anonymous categorical feature.\\
Annonymous\_c2 & Anonymous categorical feature.\\
Annonymous\_5 & Anonymous feature to describe device.\\
Annonymous\_6 & Anonymous feature to describe device.\\
Annonymous\_7 & Anonymous feature to describe device.\\\hline
\end{tabular}
\end{table}

\subsection{IP table}

This table  describes the joint behavior of device or cookie on IP address. One device or cookie may appear on multiple IPs.

\begin{table}[h!]
\begin{tabular}{|p{2.2cm}|p{6cm}|}
\hline
Feature & Description \\\hline
ID & ID of the device or cookie.\\
Device/Cookie & Boolean. If it is a device or cookie.\\
IP & IP Address.\\
Freq count & How many times have we seen the device or cookie in hte IP.\\
Count 1 & Anonymous counter.\\
Count 2 & Anonymous counter.\\
Count 3 & Anonymous counter.\\
Count 4 & Anonymous counter.\\
Count 5 & Anonymous counter.\\\hline
\end{tabular}
\end{table}

\subsection{IP Aggregation}

It provides information that describe each IP across all the devices or cookies seen on that IP.

\begin{table}[h!]
\begin{tabular}{|p{2.2cm}|p{6cm}|}
\hline
Feature & Description \\\hline
IP & IP Address.\\
Is cell & Boolean, if IP is cellular IP or not.\\
Total Freq. & Total number of observations seen on this IP.\\
Count C0 & Anonymous counter.\\
Count C1 & Anonymous counter.\\
Count C2 & Anonymous counter.\\\hline
\end{tabular}
\end{table}

\subsection{Property Observation and  Property Category}

These tables provide the information regarding websites and mobile applications that a user has visited before.  Property Observation lists the specific name of the website or mobile application and Property Category table lists the categorical information.

\begin{table}[h!]
\begin{tabular}{|p{2.2cm}|p{6cm}|}
\hline
Feature & Description \\\hline
ID & ID of device or cookie.\\
Device/Cookiel & Boolean, if it is a device or cookie.\\
Property ID & Website name for cookie and mobile app name for device.\\
Count & How many times have we seen the cookie/device on this property.\\
\hline
\end{tabular}
\end{table}

\begin{table}[h!]
\begin{tabular}{|p{2.2cm}|p{6cm}|}
\hline
Feature & Description \\\hline
Property ID & Website or mobile app identifier.\\
Property Category & Category of the website or mobile app.\\
\hline
\end{tabular}
\end{table}

\appendix[Features]
\label{features}

This section contains the description of every feature contained in the training and test sets created in section \ref{feat}.

\subsubsection{Device Features}

\begin{itemize}
\item Feature 1: Device Type
\item Feature 2: Device OS
\item Feature 3: Device Country
\item Feature 4: Device Annonymous\_c0
\item Feature 5: Device Annonymous\_c1
\item Feature 6: Device Annonymous\_c2
\item Feature 7: Device Annonymous\_5
\item Feature 8: Device Annonymous\_6
\item Feature 9: Device Annonymous\_7
\item Feature 10: Number of IP addresses associated to the Device
\item Feature 11: Number of Properties associated to the Device
\end{itemize}
\quad

\subsubsection{Cookie Features}

\begin{itemize}
\item Feature 12: Cookie Computer OS
\item Feature 13: Cookie Browser Version
\item Feature 14: Cookie Country
\item Feature 15: Device Annonymous\_c0
\item Feature 16: Device Annonymous\_c1
\item Feature 17: Device Annonymous\_c2
\item Feature 18: Device Annonymous\_5
\item Feature 19: Device Annonymous\_6
\item Feature 20: Device Annonymous\_7
\item Feature 21: Number of IP addresses visited by the Cookie
\end{itemize}
\quad

\subsubsection{Relational Features}:\\
We have extracted the following variables:
\begin{itemize}
\setlength\parskip{0.2cm}
\item $\mathbf{x}_a=\begin{bmatrix}
\mathbf{x}_{a1}\\\vdots\\\mathbf{x}_{a5}
\end{bmatrix} \in \mathbb{N}^{5 \times 1} $

Contains the aggregated information (Is Cell, Total Frequency, Count C0, Count C1 and Count C2) of the IP address a.
\item $\mathbf{z}_{ab}=\begin{bmatrix}
\mathbf{z}_{ab1}\\\vdots\\\mathbf{z}_{ab6}
\end{bmatrix} \in \mathbb{N}^{6 \times 1} $

Contains the joint behaviour (Freq Countl, Count 1, Count 2, Count 3, Count 4 and Count C5) of the device or cookie b on the IP address a.
\end{itemize}

\quad
 \\
We have also created these sets in order to extract features that represents the relation between the device and the cookie:

\begin{itemize}
\setlength\parskip{0.2cm}
\item $\mathcal{I}_{D1}$: It contains IP addresses visited by the device.
\item $\mathcal{I}_{D2}$: It contains IP addresses visited by the device that appear in less than ten devices and twenty cookies.
\item $\mathcal{I}_{C1}$: It contains IP addresses visited by the cookie.
\item $\mathcal{I}_{C2}$: It contains IP addresses visited by the cookie that appear in less than ten devices and twenty cookies.
\item $\mathcal{P}_{D}$: It contains the device's properties.
\item $\mathcal{I}   = \begin{cases} \mathcal{I}_{D1} \cap \mathcal{I}_{C1} \text{, if $\left\vert \mathcal{I}_{D2} \cap \mathcal{I}_{C2} \right\vert=0$} \\  \mathcal{I}_{D2} \cap \mathcal{I}_{C2} \text{, if $\left\vert \mathcal{I}_{D2} \cap \mathcal{I}_{C2} \right\vert>0$} \end{cases}$
\item $\mathcal{O}$: This set contains the remaining devices with the same cookie's handle.
\item $\mathcal{I}_{O}$: It contains the IP addresses visited by any device in $\mathcal{O}$.
\item $\mathcal{P}_{O}$: It contains the properties visited by any device in $\mathcal{O}$.
\end{itemize}
\quad

To create the following relational features:
\begin{itemize}
\setlength\parskip{0.2cm}
\item Feature 22: $\left\vert \mathcal{I}_{D1} \cap \mathcal{I}_{C1} \right\vert$ 
\item Feature 23: $\left\vert \mathcal{I}_{D2} \cap \mathcal{I}_{C2} \right\vert$ 
\item Feature 24: $\left\vert \mathcal{O} \right \vert$
\item Feature 25: $\left\vert \mathcal{I}_{D1}  \cap \mathcal{I}_{O} \right \vert$
\item Feature 26: $\left\vert \mathcal{P}_{D}  \cap \mathcal{P}_{O} \right \vert$
\item Features 28-32: $\displaystyle\sum_{i \in \mathcal{I}} \mathbf{x}_i$ 
\item Features 33-37: $\displaystyle\sum_{i \in \mathcal{I}}\frac{ \mathbf{x}_i}{\left\vert \mathcal{I} \right \vert}$
\item Features 38-43: $\displaystyle\sum_{a \in \mathcal{I}} \mathbf{z}_{ab}$ , where b is the device.
\item Features 44-49: $\displaystyle\sum_{a \in \mathcal{I}} \frac{\mathbf{z}_{ab}}{\left\vert \mathcal{I} \right \vert}$ , where b is the device.
\item Features 50-55: $\displaystyle\sum_{a \in \mathcal{I}} \mathbf{z}_{ab}$ , where b is the cookie.
\item Features 56-61: $\displaystyle\sum_{a \in \mathcal{I}} \frac{\mathbf{z}_{ab}}{\left\vert \mathcal{I} \right \vert}$ , where b is the cookie.
\item Features 62-67: Features 38-43 minus features 50-55.
\end{itemize}

\section*{Acknowledgment}

The author would like to thank, in the first place, the ICDM 2015: Drawbridge Cross-Device Connections Challenge organizers for their hard work, and in the second place, to everyone who participated in the challenge making it interesting and competitive.



\bibliographystyle{IEEEtran}
\bibliography{ICDMbib}
%



\end{document}